\documentclass[preprint,journal]{vgtc}         %

\ifpdf%
  \pdfoutput=1\relax                   %
  \usepackage{graphicx}                %
  \DeclareGraphicsExtensions{.pdf,.png,.jpg,.jpeg} %
\else%
  \usepackage{graphicx}                %
  \DeclareGraphicsExtensions{.eps}     %
\fi%

\graphicspath{{figures/}{pictures/}{images/}{./}} %

\usepackage{microtype}                 %
\PassOptionsToPackage{warn}{textcomp}  %
\usepackage{textcomp}                  %
\usepackage{mathptmx}                  %
\usepackage{times}                     %
\usepackage{cite}                      %
\usepackage{tabu}                      %
\usepackage{booktabs}                  %

\onlineid{1516}

\vgtccategory{Research}
\vgtcpapertype{System}

\title{OmniSyn: Synthesizing 360 Videos with Wide-baseline Panoramas}

\author{David Li$^{\dagger,\ddagger,}$\thanks{Corresponding authors: \href{mail-to:dli7319@umd.edu}{dli7319@umd.edu} and \href{mail-to:me@duruofei.com}{me@duruofei.com}}
\qquad
Yinda Zhang$^{\dagger}$
\qquad
Christian H{\"a}ne$^{\dagger}$
\qquad
Danhang Tang$^{\dagger}$
\qquad
Amitabh Varshney$^{\ddagger}$
\qquad
Ruofei Du$^{\dagger,*}$\\
\scriptsize
$^{\dagger}$ Google Research \qquad $^{\ddagger}$ University of Maryland, College Park
}

\shortauthortitle{Li \MakeLowercase{\textit{et al.}}: OmniSyn: Synthesizing 360 Videos with Wide-baseline Panoramas}

\abstract{
Immersive maps such as Google Street View and Bing Streetside provide true-to-life views with a massive collection of panoramas.
However, these panoramas are only available at sparse intervals along the path they are taken, resulting in visual discontinuities during navigation.
Prior art in view synthesis is usually built upon a set of perspective images, a pair of stereoscopic images, or a monocular image, but barely examines wide-baseline panoramas, which are widely adopted in commercial platforms to optimize bandwidth and storage usage.
In this paper, we leverage the unique characteristics of wide-baseline panoramas and present OmniSyn, a novel pipeline for 360\degree view synthesis between wide-baseline panoramas.
OmniSyn predicts omnidirectional depth maps using a spherical cost volume and a monocular skip connection, renders meshes in 360\degree images, and synthesizes intermediate views with a fusion network.
We demonstrate the effectiveness of OmniSyn via comprehensive experimental results including comparison with the state-of-the-art methods on CARLA and Matterport datasets, ablation studies, and generalization studies on street views.
We envision our work may inspire future research for this unheeded real-world task and eventually produce a smoother experience for navigating immersive maps.
}


\CCScatlist{ %
 \CCScat{K.6.1}{Management of Computing and Information Systems}%
{Project and People Management}{Life Cycle};
 \CCScat{K.7.m}{The Computing Profession}{Miscellaneous}{Ethics}
}

\vgtcinsertpkg

\usepackage{xcolor}
\usepackage{multirow}
\usepackage{makecell}
\usepackage{subcaption}
\usepackage{amsmath}
\usepackage{amsfonts}
\usepackage{amssymb}
\DeclareMathOperator{\atantwo}{arctan2}

\definecolor{ToRephaseColor}{rgb}{1.0, 0.4, 0.0}
\definecolor{TodoColor}{rgb}{1.0, 0.0, 1.0}
\definecolor{GreenColor}{rgb}{0.0, 0.0, 1.0}
\definecolor{PurpleColor}{rgb}{0.5, 0.0, 0.5}
\definecolor{BlueColor}{rgb}{0.0, 0.0, 1.0}
\definecolor{BabyBlueColor}{rgb}{0.5, 0.5, 1}
\definecolor{RevisionFixedColor}{rgb}{0.6, 0.0, 0.4}
\definecolor{FinalColor}{rgb}{0.0, 0.0, 0.0}

\newcommand{\etal}{{\em et al.}}
\newcommand{\degree}{\ensuremath{^{\circ}} }

\begin{document}

\maketitle

\section{Introduction}
Recent advances in 360\degree cameras and virtual reality headsets have promoted the interests of tourists, renters, and photographers to capture or explore 360 images on commercial platforms such as Google Street View~\cite{anguelov2010gsv}, Bing Streetside\footnote{Bing Streetside: https://microsoft.com/en-us/maps/streetside}, and Matterport\footnote{Matterport: https://matterport.com}.
These platforms allow users to virtually walk through a city or preview a floorplan by interpolating between panoramas. 
However, the existing solutions lack the visual continuity from one view to the next and suffer from ghosting artifacts caused by warping with inaccurate geometry.
While prior art reports successful view synthesis experiments in a set of perspective images \cite{broxton2020lfvideo,Chaurasia2013Depth,flynn2016deepstereo,hedman2017casual,hedman2018instant,hedman2018deepblending}, a single image \cite{kopf2020one,wiles2020synsin}, and a pair of stereoscopic panoramas with a narrow baseline \cite{attal2020matryodshka}, not much prior work addresses how we could synthesize an omnidirectional video with large movements, {\em i.e.}, using a \textit{wide-baseline} pair of panoramas.
Since wide-baseline panoramas are broadly adopted for capturing and streaming on commercial platforms, we envision view synthesis on this data can reduce the additional effort of converting to perspective images and leverage the full field-of-view for better alignment between the two panoramas.

\begin{figure}
    \centering
    \includegraphics[width=\linewidth]{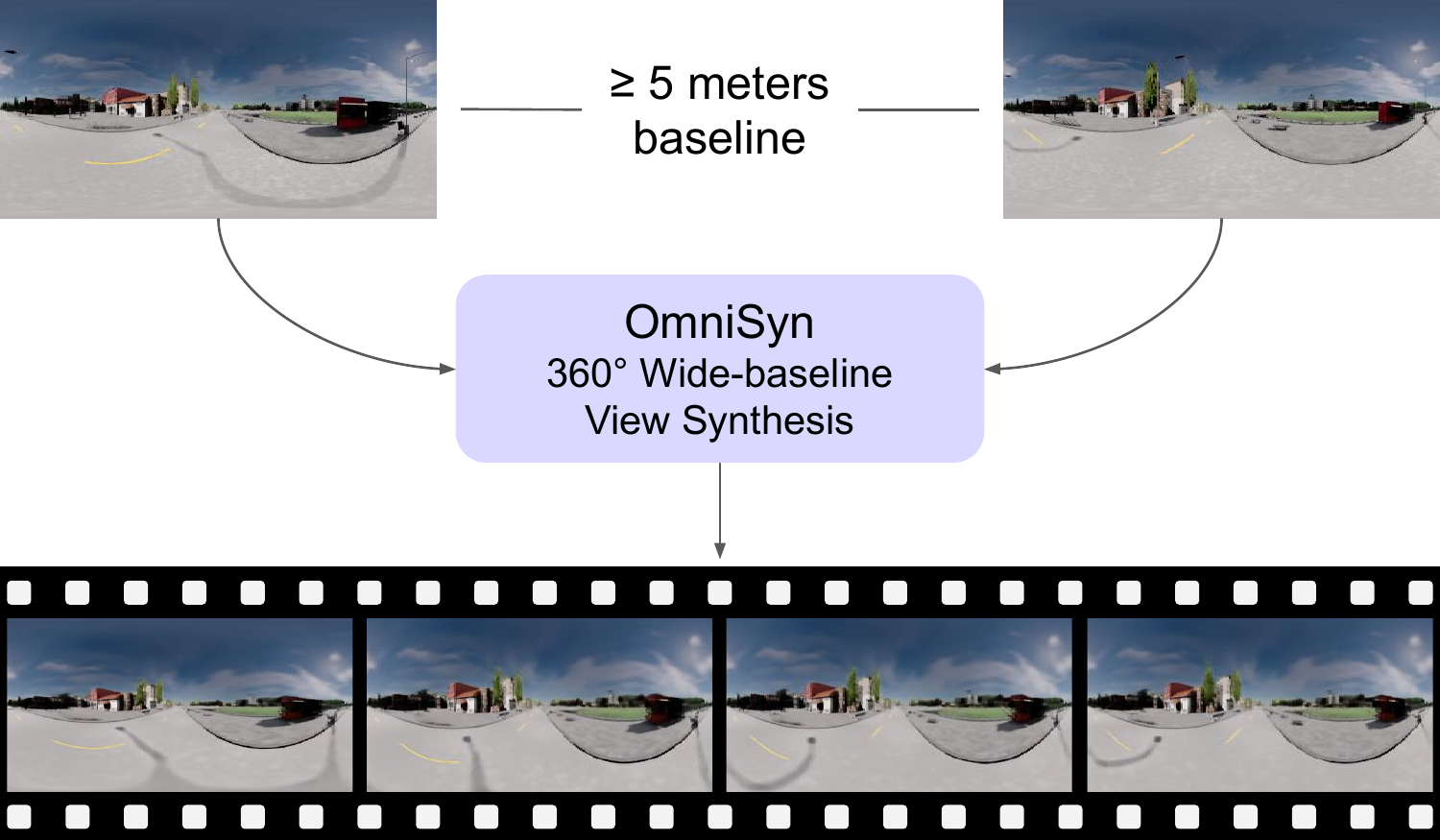}
    \caption{Given two wide-baseline 360\degree images and poses, our goal is to synthesize a video sequence of intermediate frames with plausible movement and alignment between the input images.}
    \label{fig:teaser}
\end{figure}

Our goal is to synthesize 360\degree videos between wide-baseline panoramas and stream to consumer devices for an interactive and seamless experience (\autoref{fig:teaser}).
Unlike past research which only synthesizes novel views within a limited volume \cite{broxton2020lfvideo,kopf2020one} or along a trajectory in rectilinear projection \cite{flynn2016deepstereo}, our generated 360\degree video allows users to move forward/backward, stop at any point, and look around from any perspective. This unlocks a wide range of virtual reality applications such as cinematography \cite{su2016pano2vid}, teleconferencing \cite{teo2019mixed}, and virtual tourism \cite{Du2019Geollery,Du2019Project}.

\begin{figure}
    \centering
    \includegraphics[width=\linewidth]{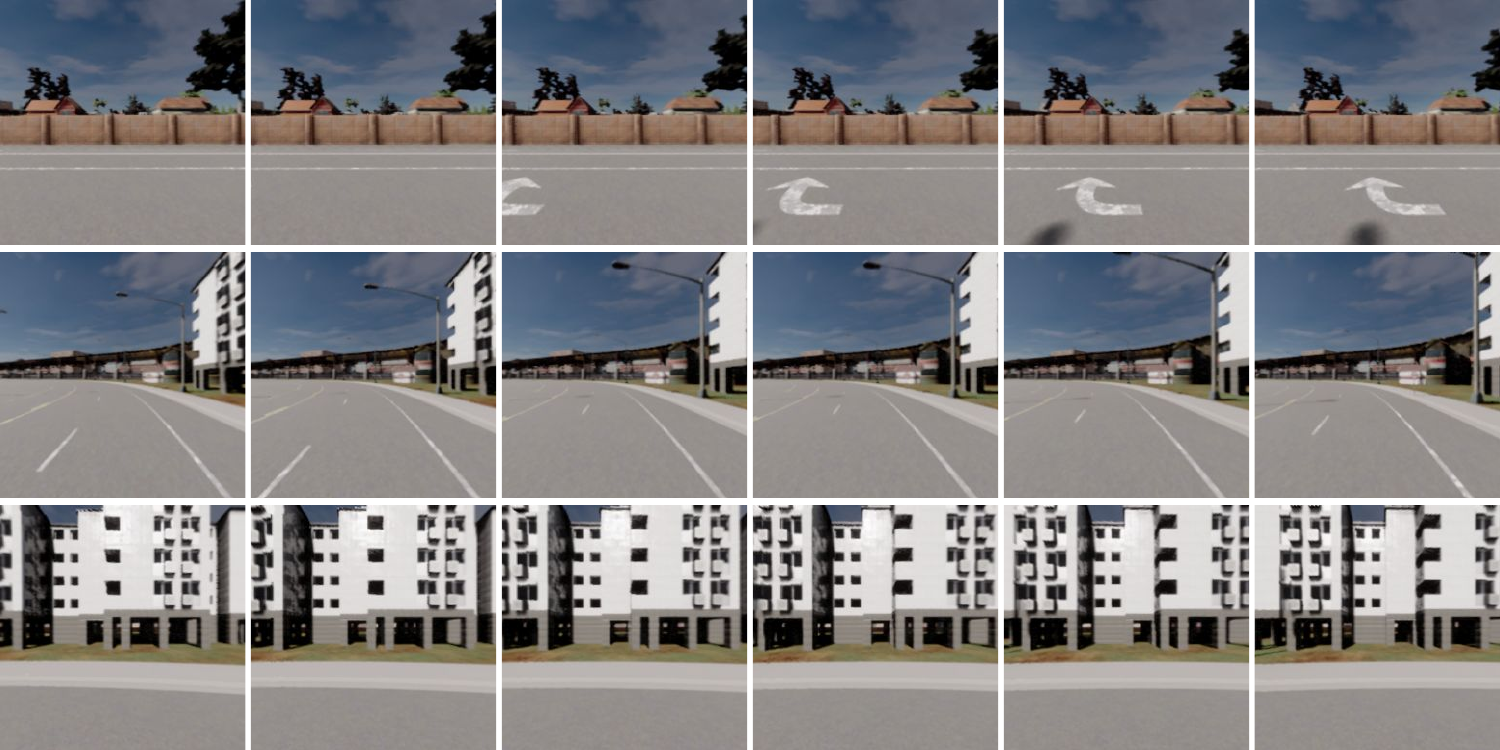}
    \caption{Examples of perspective image sequences with $1$ meter baselines. In perspective image pairs over long distances (e.g. $5$ meters), large regions of each image are not visible from the other image. Therefore existing methods cannot be applied for wide-baseline 360\degree view synthesis by simply projecting to perspective images.}
    \label{fig:perspective_examples}
\end{figure}

Classical methods for view synthesis~\cite{hedman2017casual,hedman2018deepblending} often rely on structure-from-motion~\cite{ullman1979interpretation} or multi-view stereo~\cite{hartley2003multiple} pipelines to perform a sparse 3D reconstruction and develop algorithms to densify the reconstruction.
Unfortunately, the existing approaches can hardly be applied to wide-baseline 360\degree images directly (\autoref{fig:perspective_examples}). 
On the one hand, most existing works target perspective images which encounter \textit{visual discontinuities} when objects move outside their field of view. 
On the other hand, applying monocular methods to multi-view scenarios leads to alignment issues between images as intermediate images are not fused from multiple views.
Further, real-world street view images do not have a sufficiently dense layout to apply multiview stereo methods.
So our research questions are: 
How can we achieve novel view synthesis between a pair of wide-baseline 360\degree images? 
How can we leverage the full field of view to align the pair of panoramas and inpaint occluded regions?

To answer these questions, we contribute a new pipeline for 360\degree view synthesis using \textit{wide-baseline} panoramas. Unlike the prior art, our inputs are a pair of 360\degree images which are at least $5$ meters apart for street view scenes and $2$ meters apart for indoor scenes. Our pipeline is comprised of a depth predictor, a 360\degree mesh renderer, and an image fusion network.
We train our pipeline on datasets from CARLA and Matterport and compare it with the state-of-the-art monocular view synthesis pipeline for perspective images. 
Our contributions are as:

\begin{itemize}
    \item Motivated by the goal of street view interpolation, we identify challenges associated with view synthesis between wide-baseline panoramas.
    \item We augment the classical view synthesis pipeline to address the challenges by using 360 cost volume for depth estimation, incorporating mesh rendering, and leveraging wide-baseline panoramas in the depth-estimation and fusion components.
    \item We conduct experiments against a modified version of SynSin, which is currently the state-of-the-art view synthesis pipeline. We also conduct ablation experiments to compare mesh rendering with point cloud rendering and identify the suitable scenarios for each.
\end{itemize}

\section{Related Work}
Our work builds upon a rich literature of prior art on view synthesis and neural rendering.
View synthesis has been studied in many setups such as dense image sequences, unstructured photo collections, as well as unique camera layouts. Here we provide a brief overview of the recent advances in perspective view synthesis and 360\degree view synthesis most relevant to our work.

\subsection{Perspective View Synthesis}
Existing work in view synthesis can be classified into one of the following categories: reconstruction-based, multiplane image (MPI) based, implicit-based, and warping-based methods.

Traditional methods in view synthesis reconstruct the underlying 3D geometry of the scene with multiview perspective images. Given a set of images with sufficient overlap, tools such as COLMAP \cite{schoenberger2016sfm, schoenberger2016mvs} allow sparse 3D reconstruction using structure-from-motion and multi-view stereo techniques. Such 3D reconstructions can then be used to create meshes to render views from novel perspectives \cite{fruh2003constructing,hedman2018deepblending,xiao2009image,xiao2014reconstructing}.

\begin{figure*}[!t]
    \centering
    \includegraphics[width=\linewidth]{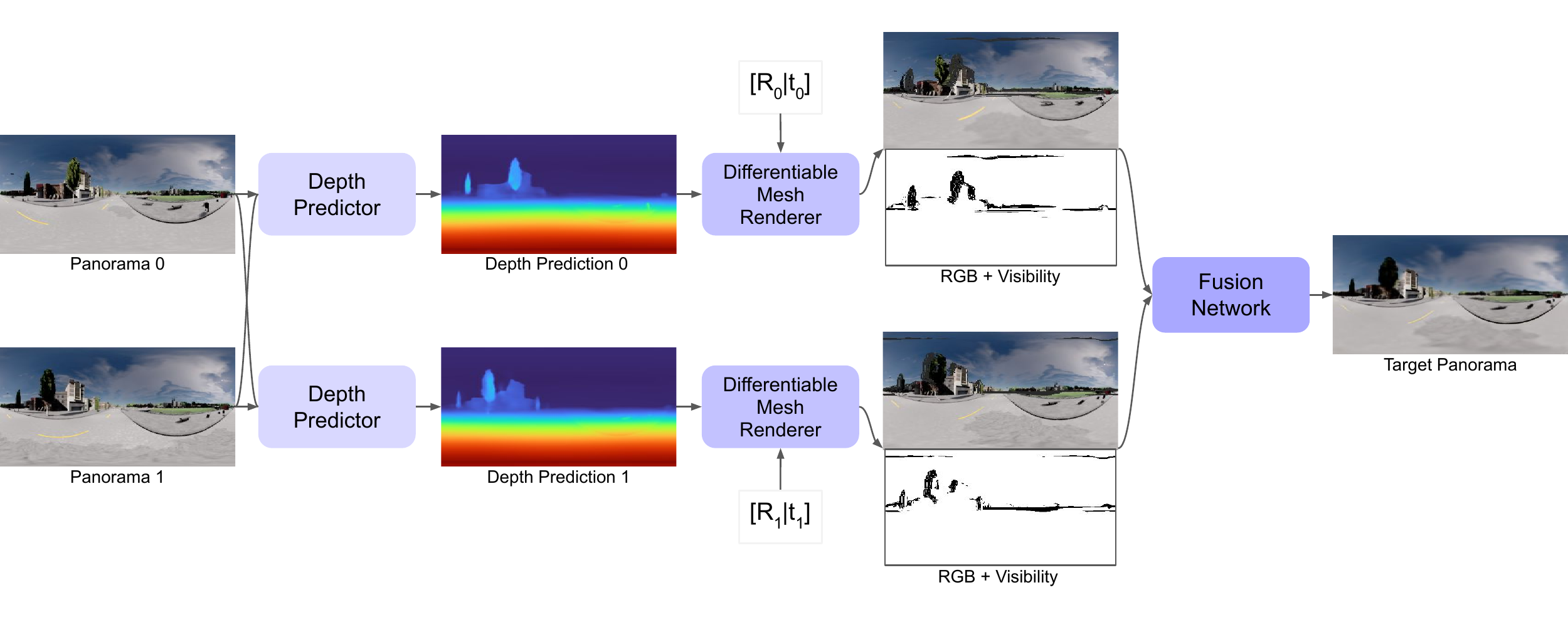}
    \caption{Our 360\degree view synthesis pipeline consists of a stereo depth predictor, a 360\degree mesh renderer, and an image fusion network. All the three components are differentiable, while only the depth predictors and image fusion network have learnable parameters.}
    \label{fig:pipeline}
\end{figure*}

\begin{figure*}[!htbp]
    \centering
    \includegraphics[width=\linewidth]{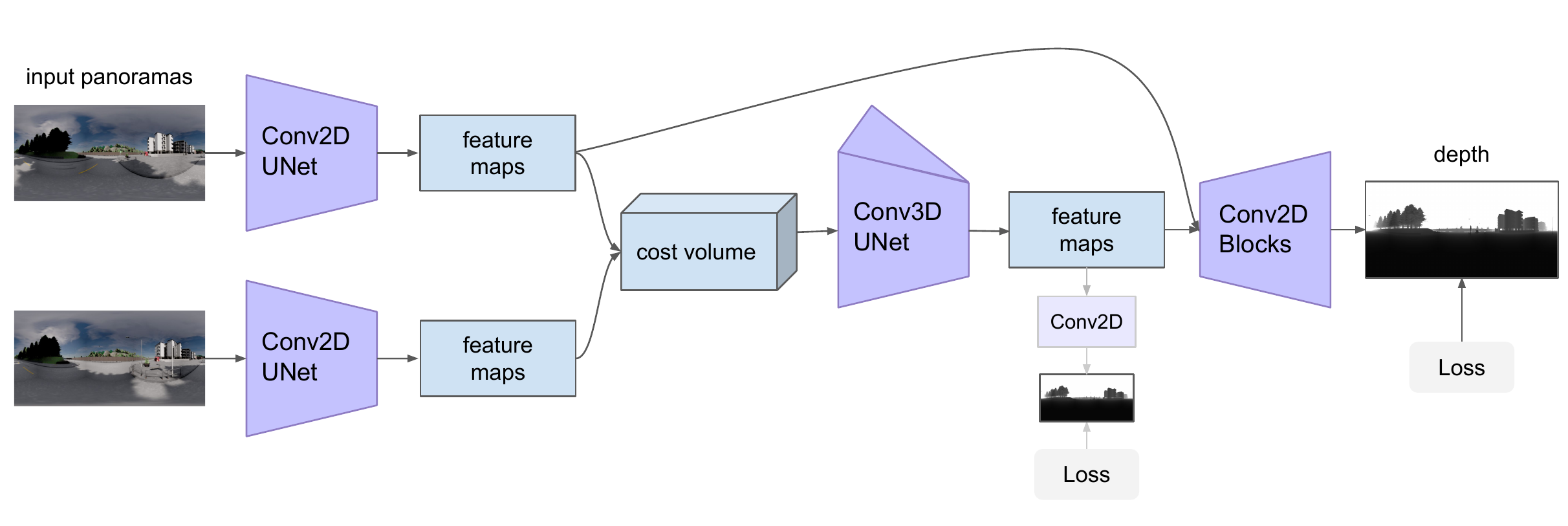}
    \caption{An overview of our stereo depth prediction network for wide-baseline 360\degree panoramas. Our depth prediction network features a stereo path which estimates depth using a spherical cost volume and a monocular skip connection which allows semantic depth estimation for occluded regions.}
    \label{fig:depth_pipeline}
\end{figure*}

Later research focuses on view synthesis using layered representations including MPI \cite{flynn2016deepstereo, flynn2019deepview, srinivasan2019pushing, tucker2020singlempi,zhou2018stereo}, layered-depth images (LDI) \cite{dhamo2019peeking, shih2020photography,tulsiani2018layer}, and multi-sphere images (MSI) \cite{attal2020matryodshka}.
Once the underlying representation is generated, these methods offer high-quality view-synthesis with standard mesh rendering. Layered methods work for both stereo inputs \cite{flynn2016deepstereo, penner2017soft}, as well as monocular methods \cite{dhamo2019peeking,tucker2020singlempi}.

Recent work on neural radiance fields (NeRF) \cite{martin2020nerfw,mildenhall2020nerf, zhang2020nerf} creates implicit representations for view synthesis from a set of perspective images. From these neural representations, views from new positions can be extracted using a volume rendering procedure. NeRF models can be trained using as few as 16 narrow-baseline images from a standard phone camera.

Warping and projection-based view synthesis are often used in low-data scenarios such as monocular and stereo view synthesis.
Chen \etal \cite{chen2019monocular} developed the transforming autoencoder which estimates the target depth and applies back-projection.
Following this work, Wiles \etal \cite{wiles2020synsin} developed SynSin which performs forward-projection via differentiable point cloud rendering. By adding a feature encoder and GAN-based decoder, SynSin is even able to properly handle occlusions synthesizing views from a single image. 

In wide-baseline view synthesis, researchers augment warping and projection with blending in developing view synthesis pipelines.
M\"uller \etal \cite{mueller2009view} perform view synthesis by masking, projecting, and filtering background and foreground layers in a 3-stage process.
Hobloss \etal \cite{hobloss2019hybrid} develop a hybrid approach for wide-baseline view synthesis with depth-based warping, two-step hole filling, and CNN-based blending procedures.

\subsection{360\degree View Synthesis}
Similar to view synthesis for perspective images, view synthesis for 360\degree images and videos has been studied with different goals and input scenarios. For instance, a 360\degree view synthesis loss can be used for self-supervised depth estimation \cite{zioulis2019spherical}.

In 360\degree view synthesis, researchers have developed pipelines for generating views from multiple cameras using traditional reconstruction pipelines.
Hedman \etal \cite{hedman2017casual} develop \textit{Casual 3D photography}, an algorithm which creates 5-DoF 360\degree scenes from 180\degree fisheye cameras. With 50 input images along a ring, they can create a dense reconstruction which allows them to perform view synthesis with depth and normal map estimations. Similar approaches have also been applied to sets of 360\degree images \cite{cho2019novel, zhao2013cube2video} and 360\degree video \cite{huang2017dof}.

Another traditional view synthesis technique involves light fields. Light fields involve capturing several images and considering each pixel as a result of a 4D plenoptic function evaluated from a light ray through free space. New images from different perspectives can then be generated by sampling from the existing set of light rays. Spherical light fields can be captured using a fisheye lens on a motorized setup~\cite{debevec2015spherical}.

360\degree view synthesis techniques have been used to create motion parallax for VR viewing \cite{attal2020matryodshka,bertel2019megaparallax, bartel2020omniphotos, luo2018parallax, serrano2019parallax}. 
Attal \etal~develop MatryODShka~\cite{attal2020matryodshka} which uses multi-sphere images to add motion parallax to ODS video.
Bertel \etal~develop OmniPhotos~\cite{bartel2020omniphotos}, a method to perform 5-DoF view synthesis from a sequence of 360\degree photos along a roughly circular path with 1-meter diameter. Layered depth image (LDI) approaches have also been extended to spheres to offer real-time 6-DoF video playback \cite{broxton2020lfvideo, lin2020deep}.

Recently, there has also been a line of research on synthesizing street view images from satellite images. Lu \etal~\cite{Lu_2020_CVPR} synthesize street view images for urban areas by predicting the depth and semantics of satelite images. Shi \etal~\cite{shi2021geometryguided} develop and end-to-end pipeline for more suburban and rural areas by predicting multi-plane images (MPIs) from satellite imagery. Li \etal \cite{li2020streetview} synthesize street-view videos by predicting a latent point-cloud representation from satellite images. 
In a closer line of work, Park \etal~\cite{park_instant_2021} combine street view images and scene models to perform image synthesis while Rafique \etal~\cite{rafique2020gaf} uses a single image.
All techniques yield impressive results but do not closely resemble the true satellite images as satellite-to-street view is an extremely under-constrained problem.

Despite the significant efforts in view synthesis, existing techniques and pipelines are unsuitable for wide-baseline 360\degree view synthesis. Many often require dense sets of images or unique camera setups and promising monocular pipelines often fail to accurately align their synthesized views with ground-truth images.
To address the challenge of wide-baseline 360\degree view synthesis from existing datasets, we have developed a pipeline tailored to this setup. Our pipeline is inspired by existing projection-based view synthesis pipelines but leverages 360\degree stereo inputs with varying baselines to estimate depth for accurate alignment between images. To enable inpainting and fusion over large baselines, we use mesh rendering which better represents the discontinuities of the underlying scene.

\section{Method}
OmniSyn, shown in \autoref{fig:pipeline}, consists of three major components to synthesize novel views for wide-baseline panoramas: a stereo depth predictor, a differentiable 360\degree mesh renderer, and an image fusion network. OmniSyn takes two wide-baseline panoramas in equirectangular projection (ERP) and poses from the street view or Matterport's metadata as input.

Given two 360\degree ERP panoramas and their relative poses to a target position, our stereo depth predictor first estimates the depth of each panorama using a spherical sweep cost volume. Based on the estimated depths, we build a mesh representation for each panorama with discontinuities computed from depth estimates. Each mesh is rendered from the target position into a separate 360\degree panorama with a corresponding visibility map. 
Following this step, our fusion network joins the two panoramas together resolving ambiguities and inpaints any holes to produce the final 360\degree panorama.

\subsection{Depth Prediction}

To perform consistent depth prediction from wide-baseline panoramas, we build a network architecture inspired by StereoNet \cite{khamis2018stereonet}, modified for stereo 360\degree depth estimation.
Our depth prediction network consists of three components: a 2D feature encoder, a 3D cost volume refinement network, and a 2D depth decoder. Stereo depth estimation allows the network to match features presented in both of the 360\degree images for aligned depth estimation while the monocular connection allows the network to predict depth for regions occluded in the other image. We follow StereoNet in first predicting a low-resolution depth. Next, instead of using the perspective cost volume, we leverage a spherical sweep \cite{sunghoon2016spherical, won2019sweepnet} cost volume of features. A refinement stage filters the cost volume and an upsampling stage guided by the feature map of the input image outputs the inverse depth prediction. %

Our depth pipeline is shown in \autoref{fig:depth_pipeline}. We use a UNet \cite{ronneberger2015u} with 5 downsampling blocks and 3 upsampling blocks for the feature encoder, a 3D UNet with 3 downsampling and 3 upsampling blocks for the cost volume refinement network, and 2 convolutional blocks for the depth decoder. Similar to 360SD-Net \cite{wang2020sdnet}, we input the vertical index as an additional channel for each convolutional layer in our depth prediction network, a technique also known as CoordConv \cite{liu2018coordconv}. This allows the convolutional layers to learn the distortion associated with the equirectangular projection (ERP). The final output convolutional layers in our depth network output one over the depth. To get the Euclidean depth, we take the inverse of this output.

\subsection{Mesh Creation}
\begin{figure*}[!h]
    \centering
    \includegraphics[width=\linewidth]{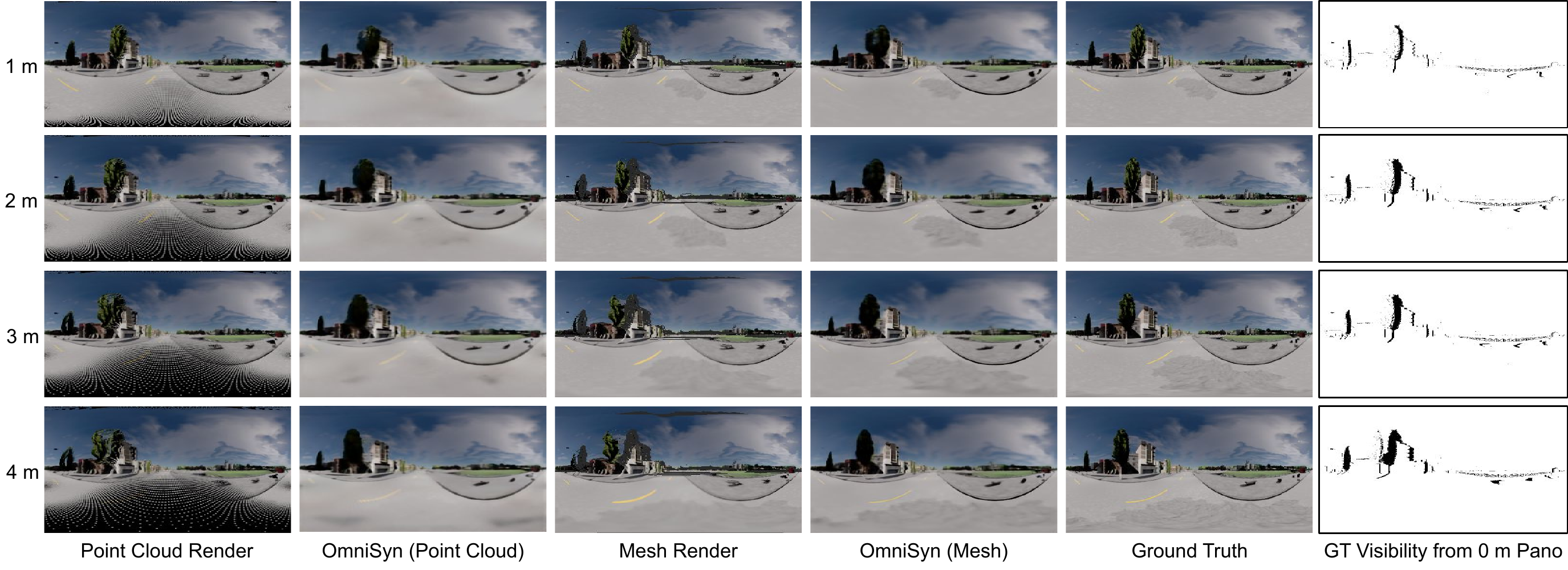}
    \caption{A comparison between mesh rendering and point cloud rendering with ERP panoramas at $1-4$ meters of movement. As moving distance increases, the distribution of nearby points in the point cloud drastically shifts, leading to sparse regions in the point cloud render and lower overall quality. In contrast, our 360\degree mesh renderer more accurately projects the input with respect to the true visibility map.
    }
    \label{fig:mesh_vs_pointcloud}
\end{figure*}

To render each image from the novel viewpoint, we first create a spherical mesh for each input image. 
By using a mesh representation rather than a point cloud representation, we avoid density issues associated with creating point clouds from ERP images, as shown in \autoref{fig:mesh_vs_pointcloud}. When moving large distances, point clouds created from ERP images contain widely varying levels of sparsity which are difficult to inpaint.

\begin{table*}[!htbp]
\begin{center}
\begin{tabular}{|l|l|c|c|c|c|c|c|c|c|c|}
\hline
Dataset & Method & IMAE $\downarrow$ & IRMSE $\downarrow$ & MAE $\downarrow$ & RMSE $\downarrow$ & $1.05$ $\uparrow$ & $1.10$ $\uparrow$ & $1.25$ $\uparrow$ & $1.25^2$ $\uparrow$ & $1.25^3$ $\uparrow$\\
\hline\hline
\multirowcell{2}[0pt][l]{CARLA\\ (ERP)} & SynSin & 0.0060 & 0.0090 & 1.19 & 5.34 & 0.8400 & 0.8837 & 0.9358 & 0.9799 & 0.9921 \\
& OmniSyn & \textbf{0.0042} & \textbf{0.0074} & 1.22 & 6.20 & \textbf{0.8473} & \textbf{0.8856} & 0.9347 & 0.9770 & 0.9913 \\
\hline
\multirowcell{2}[0pt][l]{Matterport\\ (ERP)} & SynSin & 0.1320 & 0.1775 & 0.641 & 2.119 & 0.2021 & 0.3517 & 0.6187 & 0.8838 & 0.9667 \\
& OmniSyn & \textbf{0.0518} & \textbf{0.1142} & \textbf{0.282} & \textbf{0.741} & \textbf{0.6307} & \textbf{0.7706} & \textbf{0.8936} & \textbf{0.9573} & \textbf{0.9838}\\
\hline
\end{tabular}
\end{center}
\caption{Quantitative results of Euclidean depth prediction on ERP images. Evaluating the accuracy of source depth prediction determines how well view synthesis results align between consecutive panoramas. 
For our stereo method, we present results using a 1-meter baseline between stereo images.
}
\label{table:depth_table}
\end{table*}

For a $W \times H$ resolution output image, we create a spherical mesh following the UV pattern with $2 H$ height segments and $2 W$ width segments. 
Next, vertices are offset to the correct radius based on the Euclidean depth $d$ from the depth prediction stage.
After creating the mesh and offsetting vertices to their correct depth, we calculate the gradient of the depth map along the $\theta$ and $\phi$ directions, yielding gradient images $d_{\theta}$ and $d_{\phi}$. These gradient images represent an estimate of the normal of each surface. Large gradients in the depth image correspond to edges of buildings and other structures within the RGB image. These surfaces have a normal vector perpendicular to the vector from the camera position.
To classify discontinuities in the 3D structure, we identify areas where adjacent pixels are greater than some fixed value $k$ apart.
For these areas, we discard triangles within the spherical mesh to accurately represent the underlying discontinuity. 

With the meshes created and discontinuities calculated, we use a modified version of PyTorch3D \cite{ravi2020pytorch3d} to render the mesh from the new viewpoint to a 360\degree RGBD image. The mesh renderings contain holes due to occlusions in the original images. These holes are represented in the depth image as negative values, from which we extract a visibility mask, as shown in \autoref{fig:pipeline}.

To adapt the built-in mesh renderer to output 360\degree images, we modify their rasterizer which project vertices from world-coordinates to camera-coordinates and finally to screen coordinates. Rather than multiplying vertex camera-coordinates by a projection matrix, we apply a Cartesian to spherical coordinates transformation and normalize the final coordinates to $[-1, 1]$. However, doing this results in 2 issues: triangles wrapping around the left and right edges of the ERP images may be cut-off and straight lines in Cartesian coordinates may be incorrectly mapped to straight lines in ERP image coordinates. To address the first issue, we do 2 render passes, one rotated by $180^{\circ}$, and composite the passes together so that triangles which wrap around are not missing in the final render. We address the second issue by using a dense mesh to minimize the length of each triangle in the final image. One way to address the both issues simultaneously would be to render the $6$ perspective sides of a cubemap and project the cubemap into an ERP image. However, this method incurs a significant performance and memory penalty from rendering $6$ images.

\subsection{Image Fusion}
After rendering each mesh from the new viewpoint, holes appear in each rendering due to the occlusions in the synthesized view. A naive way to fill such holes is to alpha-blend both images based on their visibility maps. However, this may still leave holes in regions occluded in both images and lead to ghosting where objects are not perfectly aligned. Thus, we use a fusion network o fuse the two mesh renderings and inpaint the holes into a single consistent panorama. Specifically, we employ a 2D UNet. We first generate a binary visibility mask to identify holes in each rendered based on the negative regions in the mesh rendering depth image. Then we input both RGB mesh renderings and the corresponding binary masks into the fusion and inpainting network to get the final fused and inpainted RGB image. 

Our fusion network is a 2D UNet with 6 downsampling blocks, 1 intermediate block, 6 upsampling blocks, and a final convolution layer. Each block consists of the following layers: Padding, Conv, LeakyReLU, Padding, Conv, and LeakyReLU. Downsampling is performed using average pooling at the end of each downsampling block. Upsampling is performed using bilinear interpolation at the beginning of each block. We use circular padding at each convolutional layer, simulating Circular CNNs \cite{schubert2019circular}, to join the left and right edges. The top and bottom of each feature map use zero padding. The inputs to our network are mesh renderings and visibility maps from each input panorama, totalling eight channels. The output is three channels of RGB representing the final inpainted frame.

\subsection{Training}
\subsubsection{Loss Functions}
Our network is end-to-end differentiable but training can also be performed in two supervised stages to increase modularity. In our experiments, we train each stage for $48$ hours for a total of $96$ hours of training on a single GPU. In the first stage, we train the stereo depth predictor. Our inputs for the depth predictor are two wide-baseline images in a sequence. The depth predictor includes two heads, an intermediate which predicts a low-resolution depth $d_{pred\_low}$ with $n_{low}$ pixels based on only the cost volume and a final head which predicts a higher resolution depth $d_{pred\_hi}$ with $n_{hi}$ pixels from the feature map and the cost volume. The intermediate head is used to ensure that gradients flow through the 3D cost volume UNet. Our loss function for depth is:
\begin{equation}
\ell_{depth} = \frac{1}{n_{hi}}\left\Vert \frac{1}{d_{gt}} - \frac{1}{d_{pred\_hi}} \right\Vert_1 + \frac{\lambda}{n_{low}} \left\Vert \frac{1}{d_{gt}} - \frac{1}{d_{pred\_low}} \right\Vert_1
\end{equation}
For our experiments, we use $\lambda = 0.5$.

In the second stage, we train the fusion network using sequences of 3 panoramas: $(p_0, p_1, p_2)$. Mesh renders are generated from the first and last panoramas $(p_0, p_2)$ using the pose of the intermediate panorama $p_1$. The fusion network receives these mesh renders and combines them to predict the full intermediate panorama $p_{pred}$. The ground-truth intermediate panorama $p_1$ is used for supervision. As the $l_1$ loss has been found to perform better for image-to-image applications~\cite{zhao2017loss}, we use it to supervise the fusion network output:
\begin{equation}
\ell_{fusion} = \left\Vert p_1 - p_{pred} \right\Vert_1 
\end{equation}

Our total loss is:
\begin{equation}
\ell_{total} = \ell_{depth} + \ell_{fusion}
\end{equation}

\subsubsection{Linearization of Arccos}
Oftentimes in our pipeline, we need to convert between Cartesian and spherical coordinates for operations such as  performing transformations or rasterizing to ERP images.
The standard equations for converting from Cartesian to spherical coordinates are:
\begin{align}
\theta &= \atantwo(z, x)\\
\phi &= \arccos\left(\frac{y}{\sqrt{x^2 + y^2 + z^2}}\right)\\
r &= \sqrt{x^2 + y^2 + z^2}
\end{align}
However, the derivative of $\arccos(\tilde{y})$ is $\frac{-1}{\sqrt{1-\tilde{y}^2}}$. Near the north and south poles, $\tilde{y}^2 = \frac{y^2}{x^2 + y^2 + z^2} \approx 1$, which leads to numerical issues computing the gradient in the backwards pass.
To address this issue, we use a linear approximation for $\arccos$ for all points within $\alpha$ degrees of the north and south poles. This mitigates numerical issues by clipping the maximum and minimum gradient values near the poles, but preserving the sign of the gradient. In our experiments, we found that using $\alpha = 10$ degrees ($\approx 0.174\text{ rad}$) is sufficient for avoiding numerical issues in the backwards pass.
During training, we use the following formula for $\phi$:
\begin{equation}
\phi = 
\begin{cases}
\arccos\left(\frac{y}{r}\right) & \text{if } \left|\frac{y}{r}\right| < \cos(\alpha) \\
\alpha \cdot \frac{1 - y/r}{1 - \cos(\alpha)} & \text{if } y > 0 \text{ and } \left|\frac{y}{r}\right| \geq \cos(\alpha) \\
\pi - \alpha  \cdot \frac{1 + y/r}{1 - \cos(\alpha)} & \text{if } y \leq 0 \text{ and } \left|\frac{y}{r}\right| \geq \cos(\alpha)
\end{cases}
\end{equation}

\begin{figure*}[!htbp]
    \centering
    \includegraphics[width=0.9\linewidth]{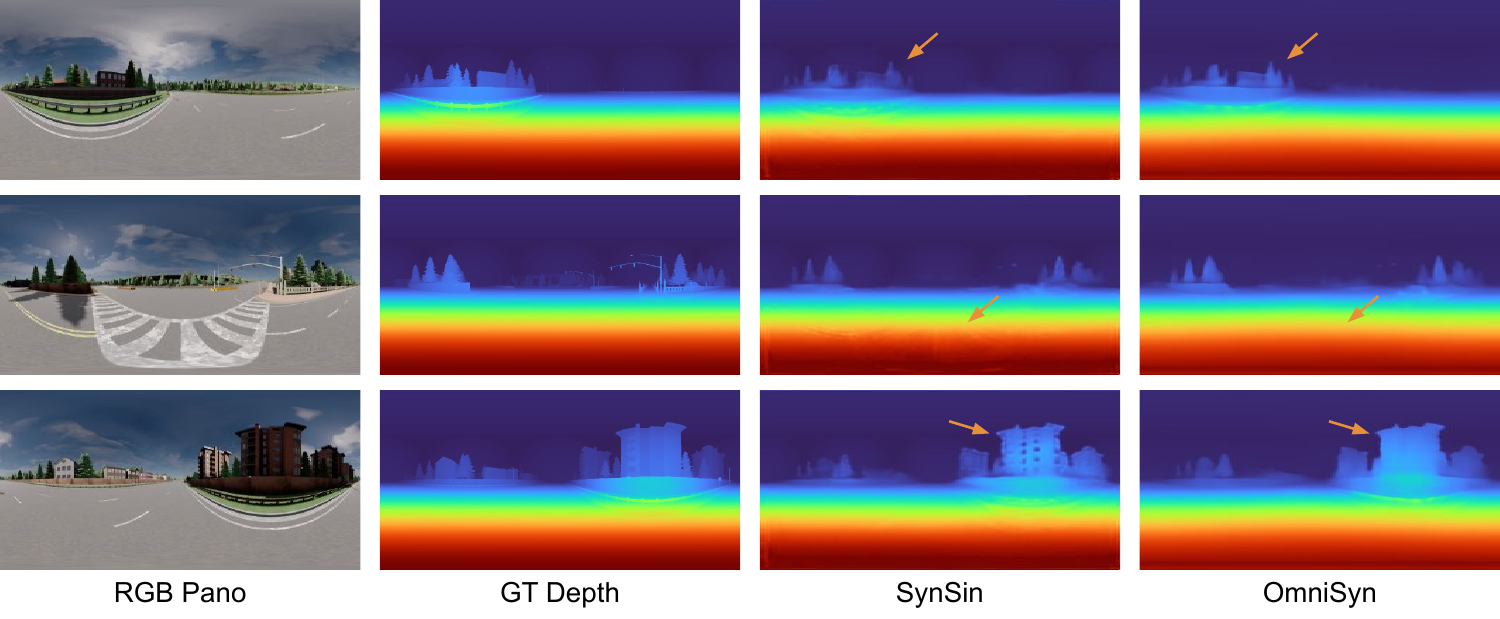}
    \caption{Qualitative results for Euclidean depth prediction on the CARLA dataset.}
    \label{fig:carla_qual_depth}
\end{figure*}

\begin{figure}[!htb]
    \centering
    \begin{subfigure}{0.46\linewidth}
      \centering
      \includegraphics[width=\linewidth]{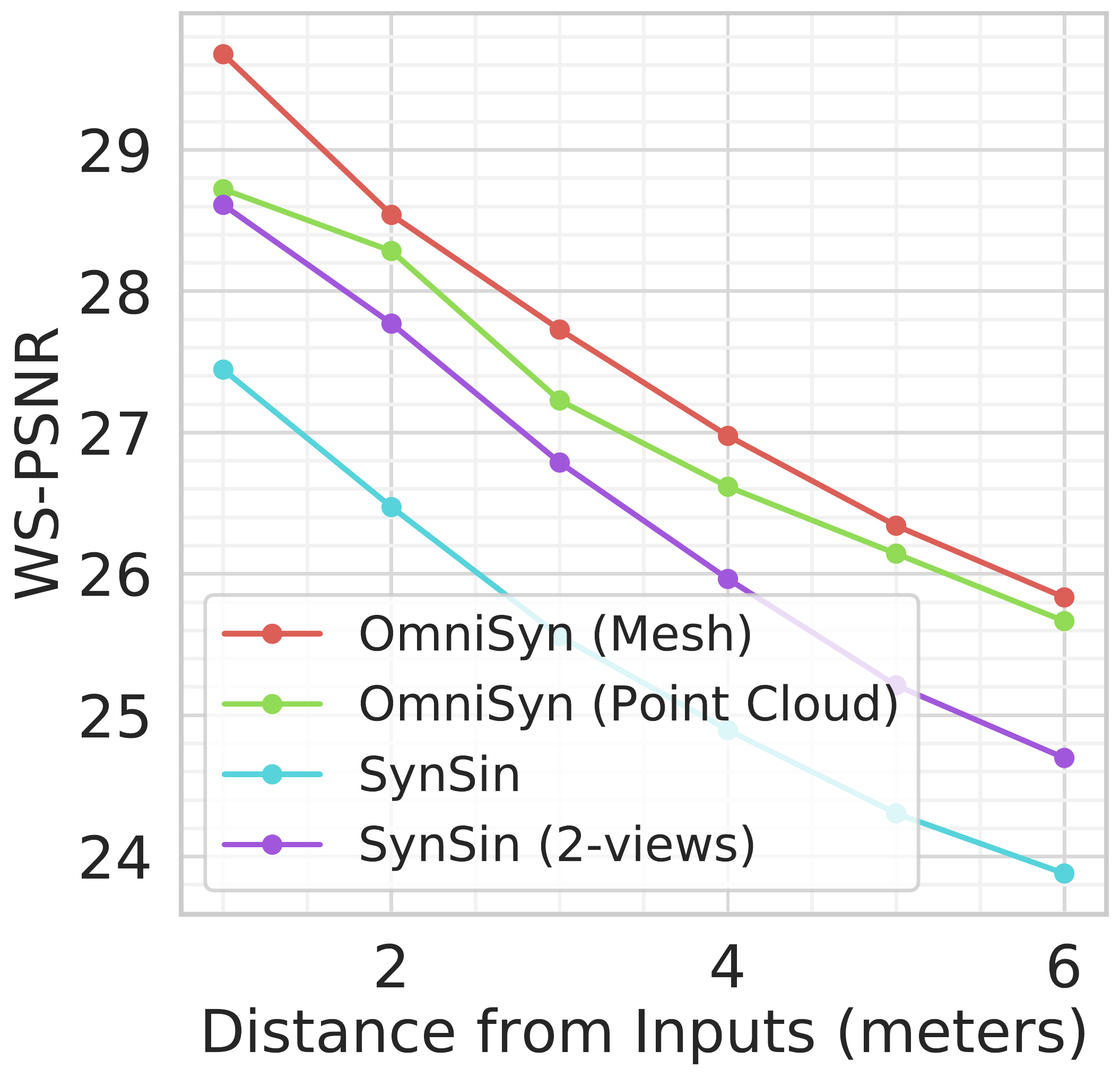}
      \includegraphics[width=\linewidth]{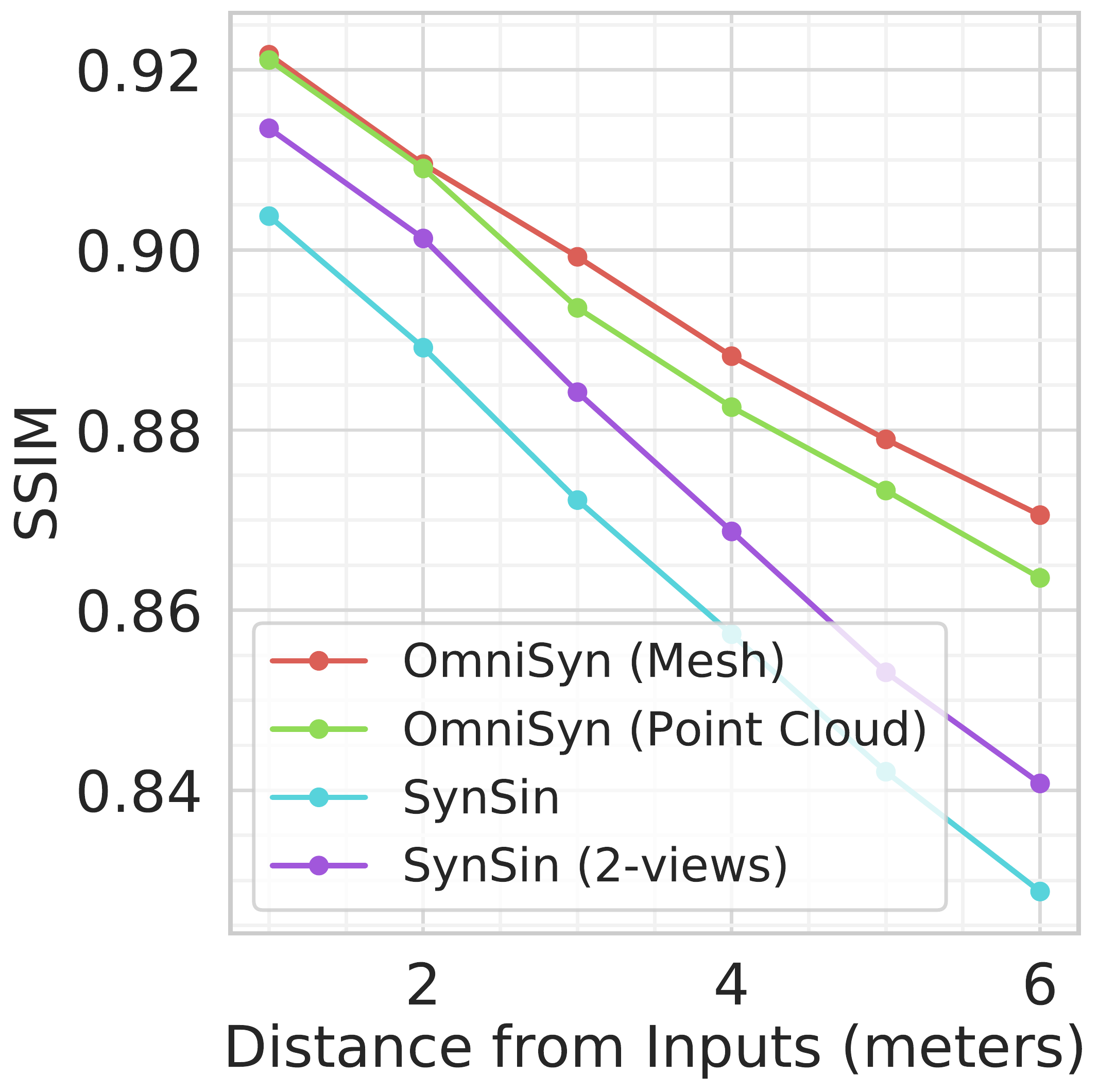}
      \includegraphics[width=\linewidth]{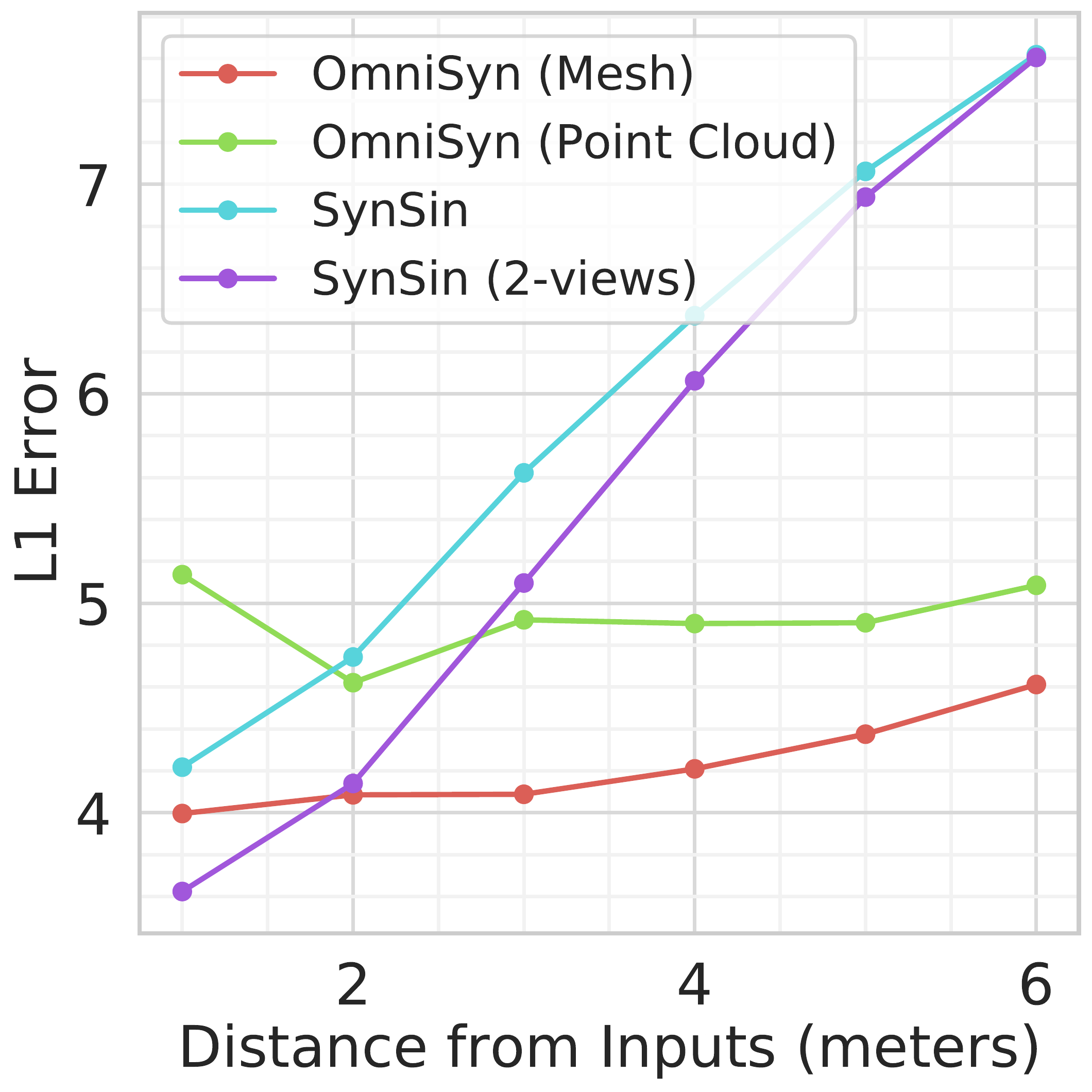}
      \caption{Carla Synthesis Quality}
      \label{fig:wspsnr_vs_distance_carla}
    \end{subfigure}
    \begin{subfigure}{0.46\linewidth}
      \centering
      \includegraphics[width=\linewidth]{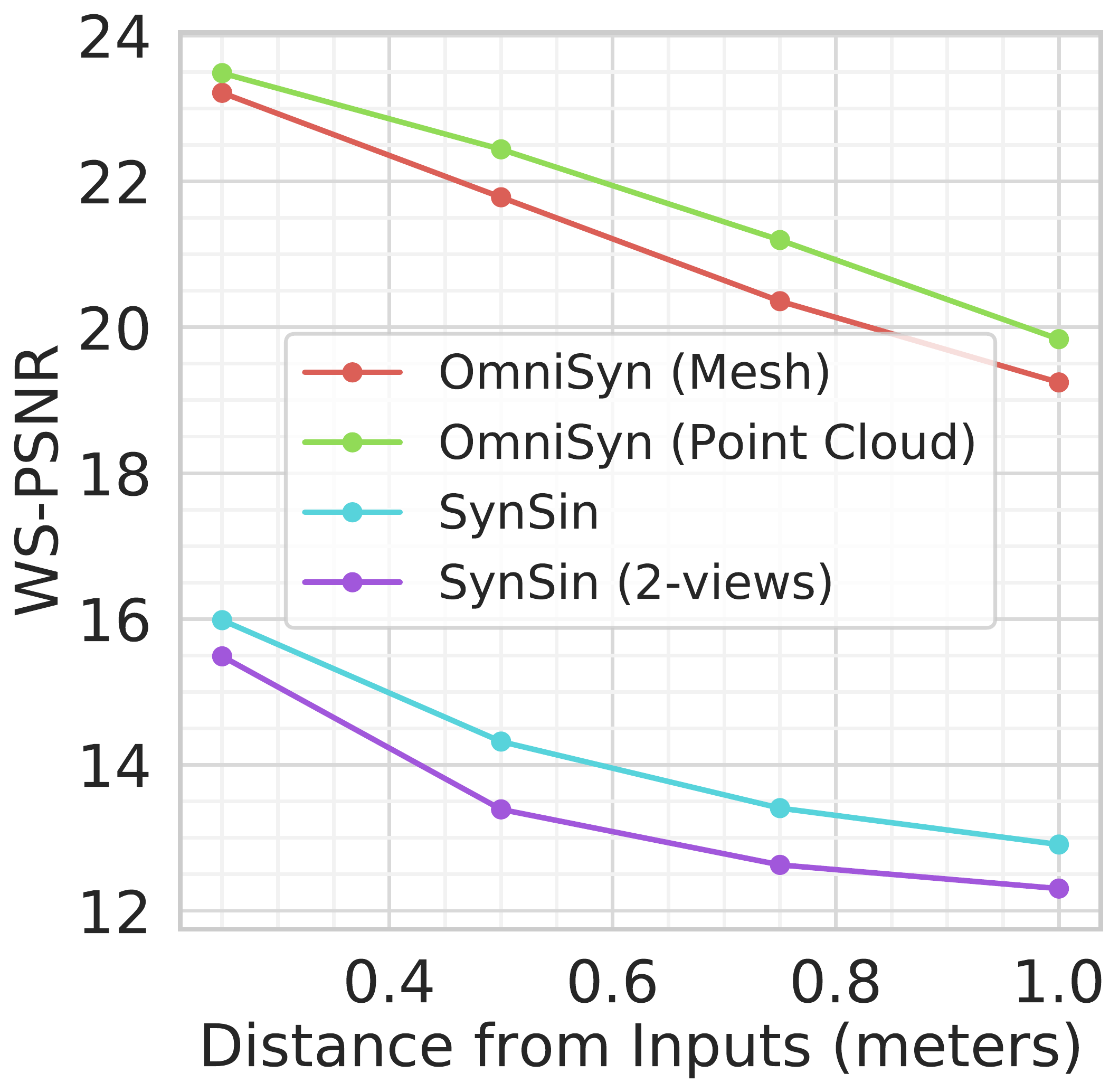}
      \includegraphics[width=\linewidth]{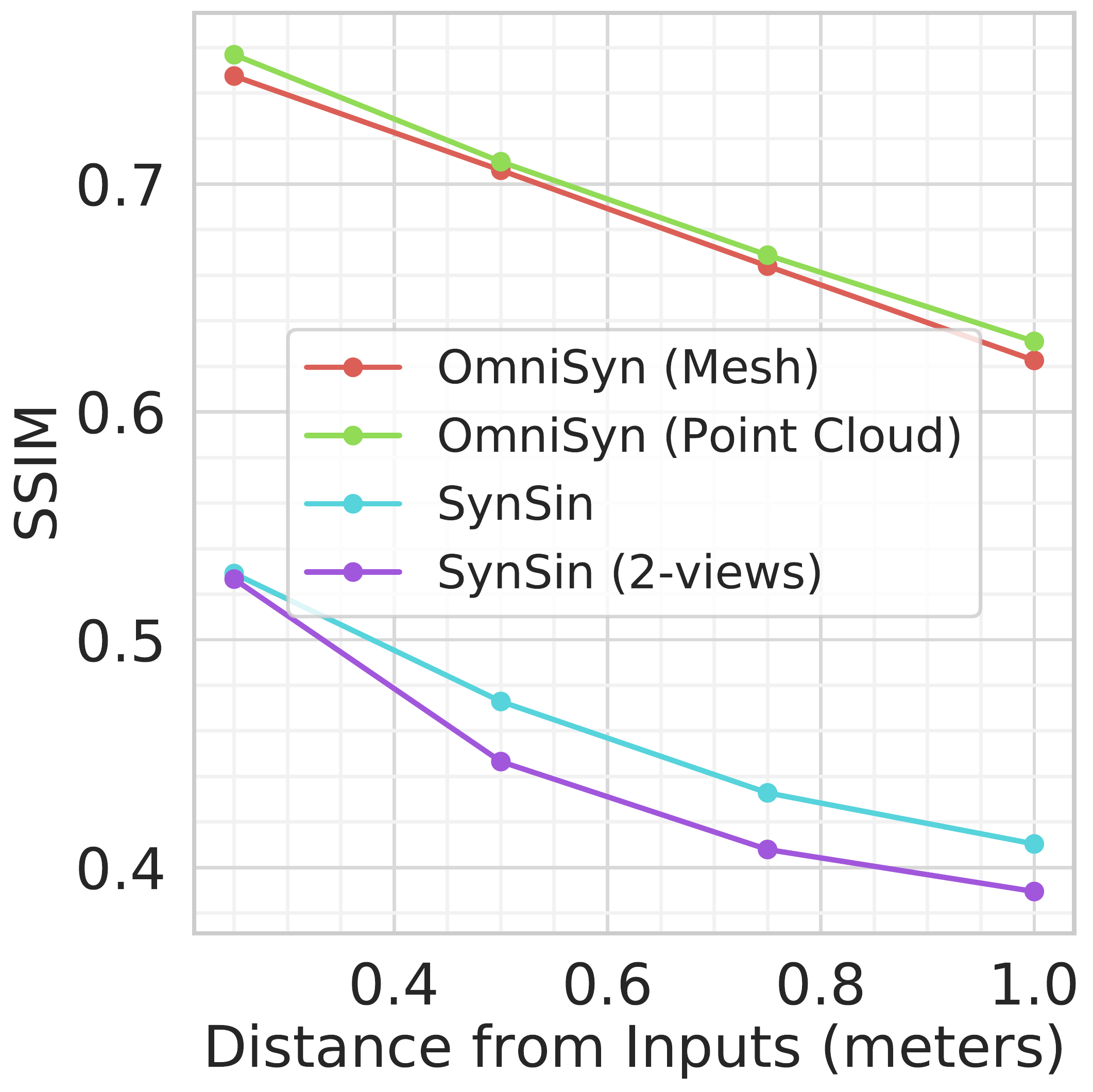}
      \includegraphics[width=\linewidth]{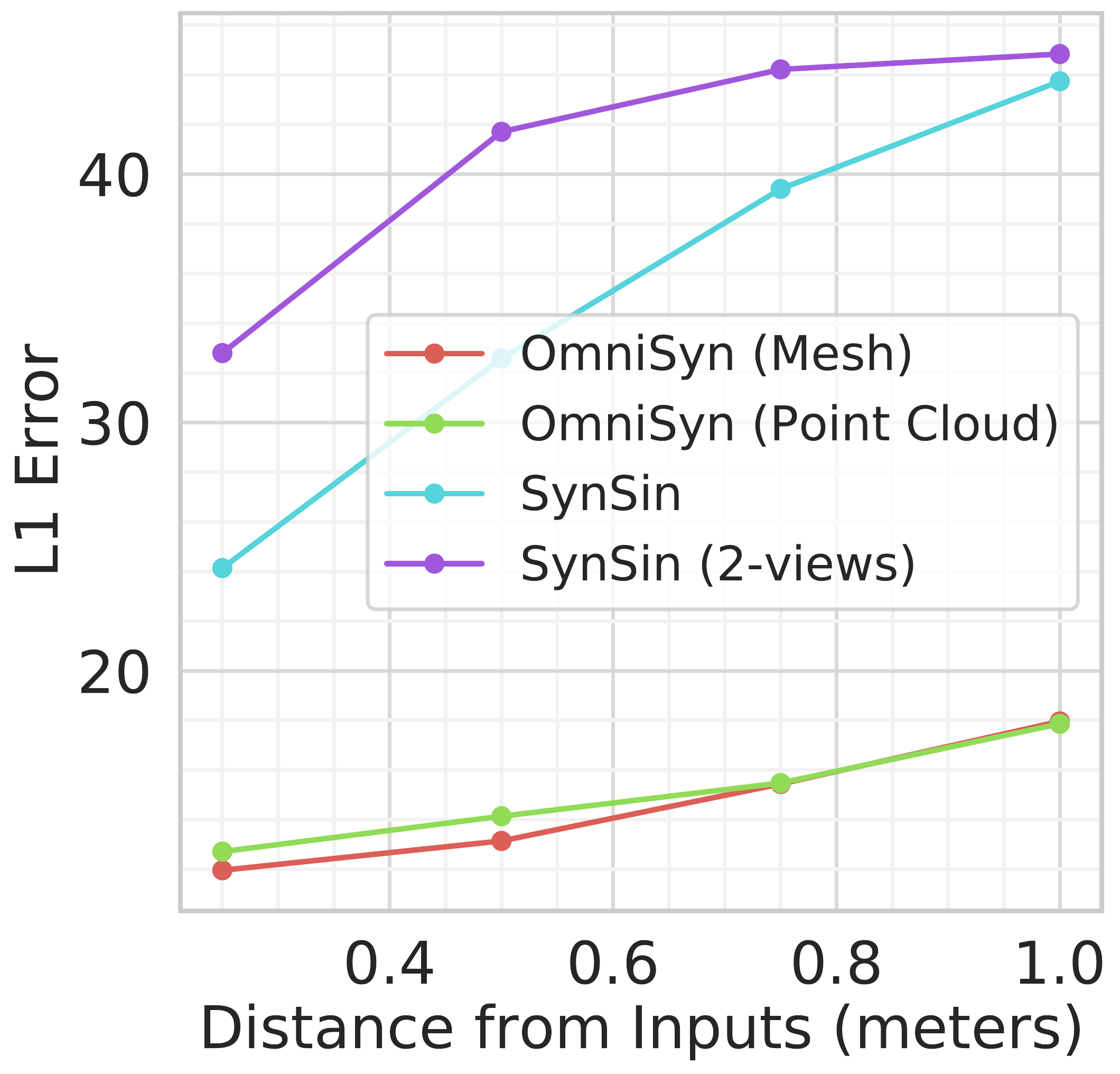}
      \caption{Matterport Synthesis Quality}
      \label{fig:wspsnr_vs_distance_m3d}
    \end{subfigure}%
    \caption{Quantitative view synthesis results of OmniSyn on Carla and Matterport3D datasets. 
    For $360$\degree view synthesis, OmniSyn outperforms SynSin for both indoor and outdoor scenes. However, there is a quality trade-off associated with using mesh over point clouds in indoor scenes.
    }
    \label{fig:wspsnr_vs_distance}
\end{figure}

\begin{figure*}[!ht]
    \centering
    \includegraphics[width=\linewidth]{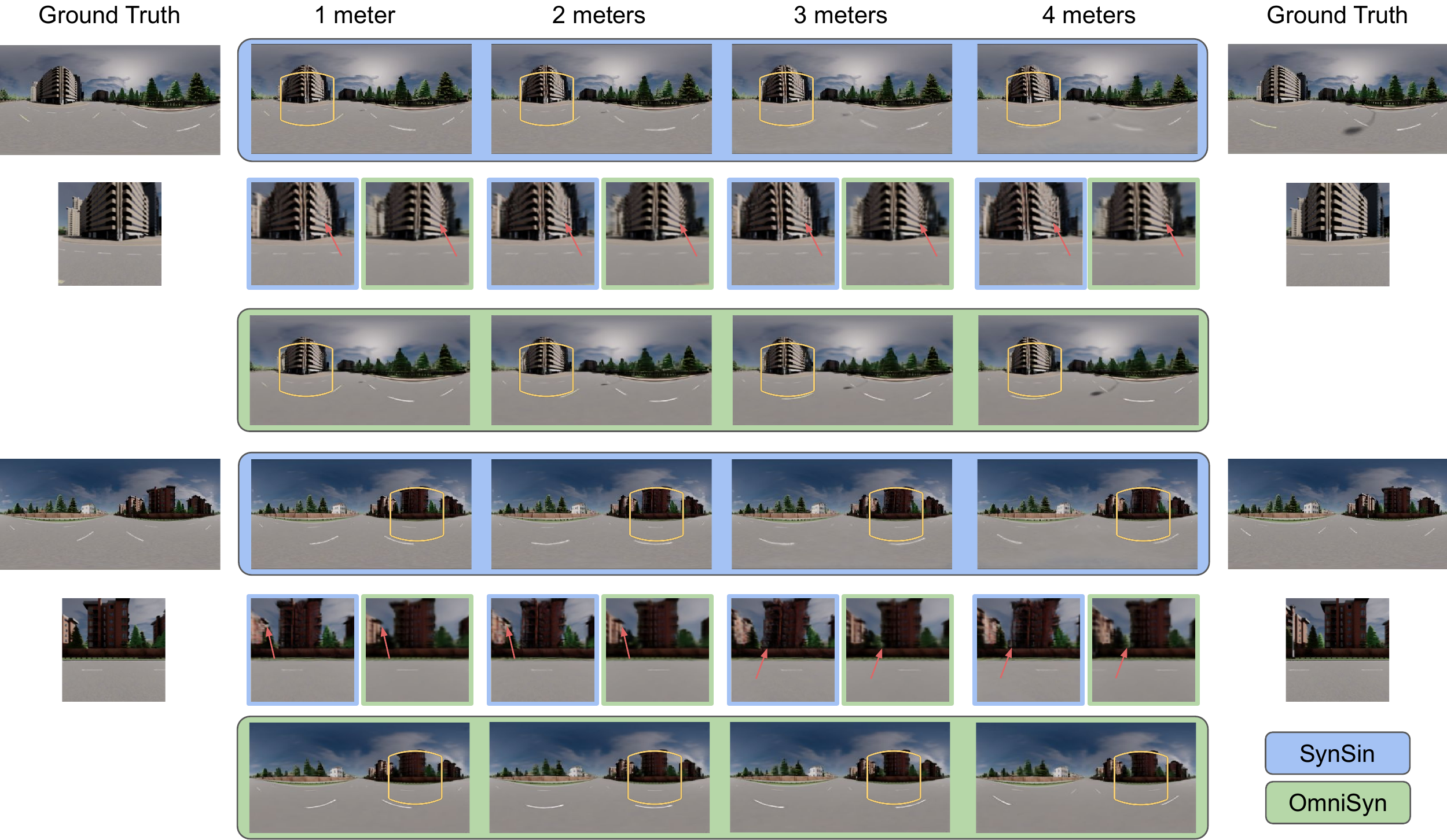}
    \caption{Qualitative results of 360\degree view synthesis on outdoor scenes of the CARLA dataset with inputs $5$ meters apart. By leveraging stereo panoramas for depth prediction and synthesis, our method is able to generate views to accurate metric scale distance and render sides of buildings that may be unclear or less sharp when generated from only a single view.
    }
    \label{fig:carla_qualitative}
\end{figure*}

\section{Experiments}
We conduct experiments comparing our pipeline to the state-of-the-art view synthesis pipeline, SynSin \cite{wiles2020synsin}. Our experiments are conducted on synthetic outdoor scenes from CARLA \cite{dosovitskiy17carla} and real-world indoor scenes from Matterport3D \cite{chang2017matterport}. To evaluate how models from each pipeline adapt to variable wide-baseline scenarios, our evaluations focus on view-synthesis results across variable distances and source depth accuracy which is essential for alignment between images synthesized from consecutive images.

\subsection{Datasets and Metrics}
We conduct experiments on synthetic street scenes from CARLA \cite{dosovitskiy17carla} as well as real-world indoor scenes from Matterport3D \cite{chang2017matterport}.
For each dataset of scenes, we compare source depth prediction and view synthesis results. Source depth accuracy provides insight into how well view synthesis results align with adjacent panoramas. View synthesis results provide insight into the accuracy of inpainting and fusion.

For CARLA, we use scenes from Town 1 to Town 6. Towns 1, 2, 3, and 4 are used as training towns with Town 5 used for validation. Town 6 is used for testing. We customize each scene to normalize weather conditions and to remove all cars and pedestrians. For each town, we generate sequences of $1000$ frames from each of the first $40$ starting points. Each frame consists of an RGB panorama, depth panorama, and pose metadata. Panoramas are created by rendering the $6$ sides of cubemaps at $256 \times 256$ for each frame and stitching them into $1024 \times 512$ ERP images. During training, we resize and subsample each sequence of frames to create $256 \times 256$ image sequences with a fixed distance between frames. We use baselines from $1$ meter to $6$ meters with $1$ meter intervals during training. For illustrative purposes, we display results with a $2:1$ aspect ratio by resizing the $256 \times 256$ outputs.

For Matterport3D, we use Habitat-Sim \cite{habitat19iccv} to generate frames at training and inference time. Similar to CARLA, we generate $6$ $256\times 256$ perspective images representing cubemap sides and stitch them into a single $256\times 256$ ERP image. Our train and test split follows that of SynSin. During training and evaluation, we use sequences of $3$ panoramas with a fixed baseline from $0.25$ meters to $1$ meter with $0.25$ meter intervals. The middle panorama is synthesized based on the first and last panorama in the sequence.

\begin{figure*}[!htbp]
    \centering
    \includegraphics[width=\linewidth]{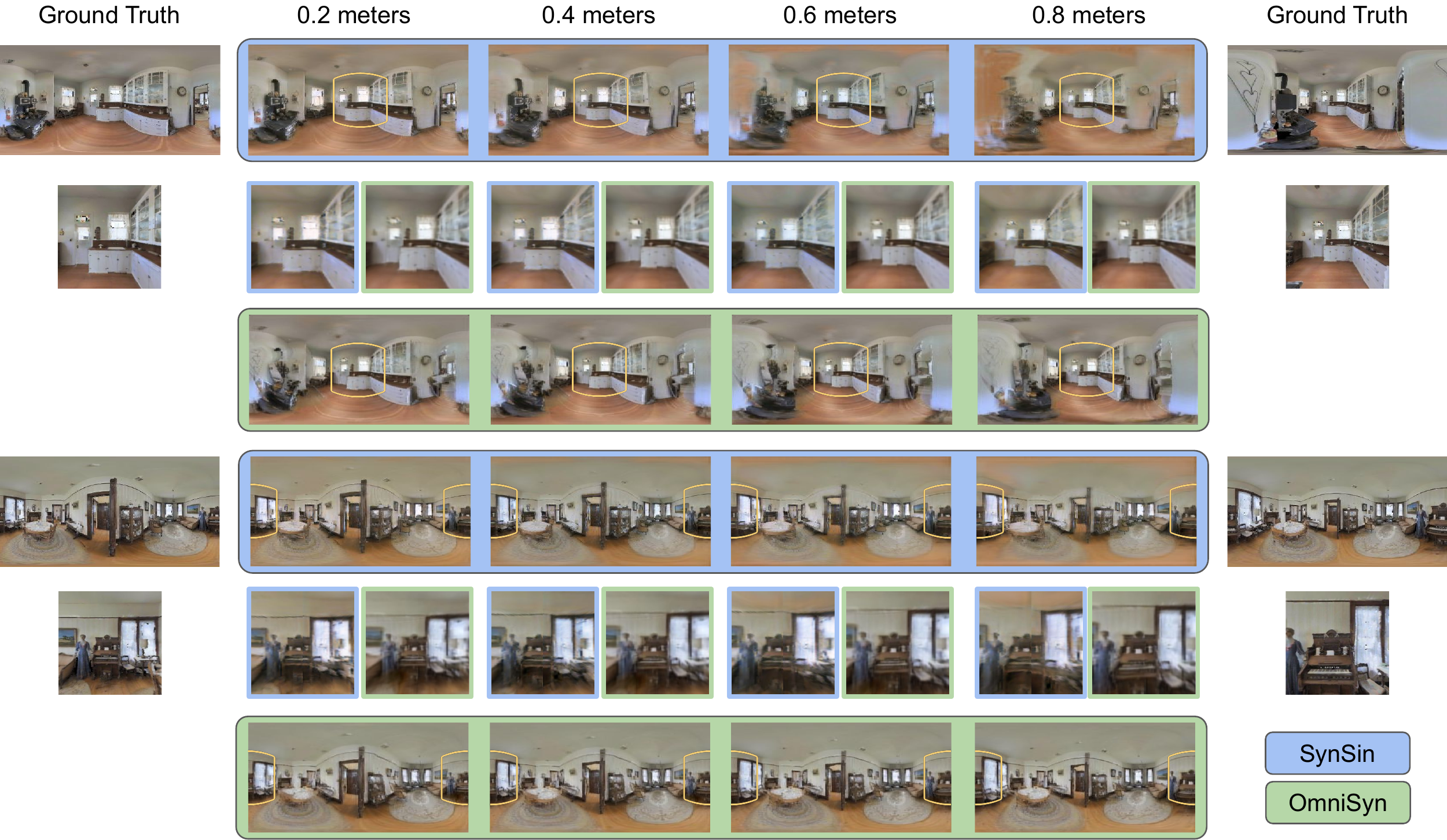}
    \caption{Qualitative results of 360\degree view synthesis on indoor scenes of the Matterport3D dataset. The pair of ground truth images are spaced $1$ meter apart. 
    }
    \label{fig:m3d_qualitative}
\end{figure*}

For comparing the accuracy of Euclidean source depth prediction, we use mean absolute error of the inverse depth (IMAE) and root mean squared error of the inverse depth (IRMSE) following the KITTI Depth Completion Evaluation benchmark\footnote{KITTI benchmark: \url{http://cvlibs.net/datasets/kitti/eval_depth.php?benchmark=depth_completion}}. Inverse depth weights errors in closer depths greater than errors in further depths and is directly proportional to disparity in perspective images. We also present the standard mean absolute error (MAE) and root mean squared error (RMSE) results. Inverse depth metrics are measured in one over meters and standard depth metrics are measured in meters. Following existing depth-estimation papers \cite{zhang2018deep}, we present $\delta < [1.05, 1.10, 1.25, 1.25^2, 1.25^3]$. To omit outliers and missing regions in each dataset, such as the sky, we compute results with respect to valid depth regions in each dataset between $1$ and $50$ meters. Our full depth prediction results are shown in \autoref{table:depth_table}. Qualitative results for depth prediction are shown in \autoref{fig:carla_qual_depth} using the Turbo colormap\footnote{Turbo colormap: \url{https://ai.googleblog.com/2019/08/turbo-improved-rainbow-colormap-for.html}}.

To compare view synthesis results for spherical ERP images, we use Weighted-to-Spherically-uniform PSNR (WS-PSNR) \cite{sun2017wspsnr}. View synthesis results are computed for a variety of distances for both datasets. We also provide results of qualitative comparison in \autoref{fig:carla_qualitative} and \autoref{fig:m3d_qualitative} on both CARLA and Matterport3D datasets.

\subsection{Comparison to SOTA}
We compare the depth and view synthesis results of our method to that of the current state-of-the-art view synthesis method, SynSin \cite{wiles2020synsin}, which also uses geometric based warping.
To adapt SynSin which is designed for perspective images to 360\degree images, we modify their perspective point cloud renderer to project 360\degree point clouds and render 360\degree ERP images. 

While SynSin is designed to work with only one image, we also compare with a 2-view version of SynSin in \autoref{fig:wspsnr_vs_distance}. For the 2-view version of SynSin, the depth predictor and feature encoder operates independently for each input image. Then both latent point clouds are combined by concatenating the list of points. A single output image is generated by decoding a single feature map from the combined point cloud.

For both SynSin and OmniSyn, we train in a supervised manner using ground-truth depth from both datasets. While SynSin provides compelling results with small movements, OmniSyn performs better in aligning with the second ground truth image and inpainting occluded areas with large movements.

\subsection{Robustness to Wide Baseline}
To determine how different 3D representations respond to varying baselines of inputs, we conduct an ablation study comparing the results of point-cloud rendering and mesh rendering. As shown in \autoref{fig:mesh_vs_pointcloud}, point cloud representations benefit from small movements as discontinuities do not need to be explicitly calculated based on normal estimates. However, for large movements needed in wide-baseline view synthesis, point clouds suffer from sparsity in their representation, leading to worse inpainting results than the mesh renderer. Quantitative results are shown in \autoref{fig:wspsnr_vs_distance}.

\begin{figure}[!htbp]
    \centering
    \includegraphics[width=\linewidth]{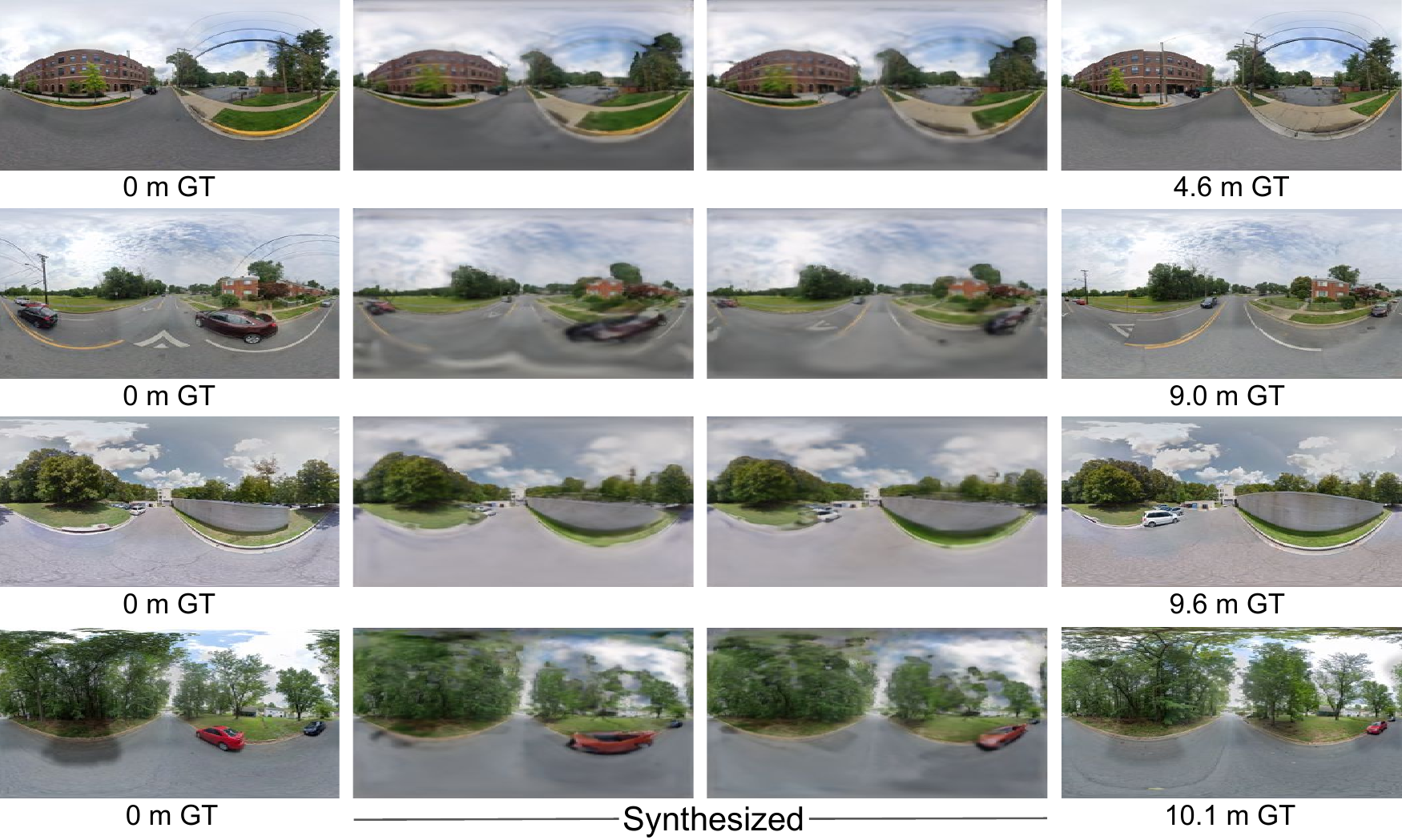}
    \caption{Generalization of our model trained on the CARLA dataset to suburban street view images.}
    \label{fig:gsv}
\end{figure}

\subsection{Generalization to Real Street View Images}
To qualitatively evaluate how well our model generalizes to real street view images, we run our stereo model over selected images from Google Street View \cite{anguelov2010gsv}. We select suburban scenes without moving cars or pedestrians to match the static scene scenario of the CARLA training dataset. For each set of input images, we generate two intermediate images using an estimated pose from geographic (latitude, longitude, heading) metadata. Qualitative results are shown in \autoref{fig:gsv}.

\section{Discussion}
We have shown that OmniSyn outperforms the state-of-the-art monocular view synthesis pipeline on CARLA and Matterport datasets. While our pipeline requires a pair of wide-baseline panoramas, such data is widely available on commercial platforms such as Google Street View and Bing Streetside. We discuss our key findings and limitations.

\subsection{Observations}

OmniSyn handles occlusions using stereo inputs and a fusion network that fuses stereo inputs and inpaints occluded regions in both images whereas SynSin handles occlusions using latent features and a GAN-based decoder network. When comparing quantitative results from SynSin and OmniSyn, we see that stereo inputs outperform GAN-based feature decoding. This is especially true for indoor scenes as shown in \autoref{fig:m3d_qualitative}.

In the case of indoor scenes from Matterport3D, monocular depth prediction yields less accurate depths than stereo depth prediction as shown in \autoref{table:depth_table} due to the greater variability of scenes. This makes monocular methods unsuitable for wide-baseline view synthesis from stereo panoramas where accurate depth is crucial for aligned view synthesis. We see in \autoref{fig:wspsnr_vs_distance_m3d}, the 2-view version of SynSin actually performs slightly worse on the indoor scenes due to the inaccurate monocular depth. However, for outdoor scenes from Carla where the depth of the road is fairly consistent between all images, monocular depth prediction yields similar results to our stereo depth. This allows OmniSyn and the 2-view version of SynSin to both outperform the single-view SynSin.

Comparing mesh rendering and point cloud renderings, we see in \autoref{fig:mesh_vs_pointcloud} that point cloud rendering has high sparsity in regions closer to the camera, forcing the inpainting or fusion networks to fill-in more of the road. In scenarios where there are distinct textures on the road, such as lane markers and traffic symbols, this causes OmniSyn to generate incorrect textures. When looking at quantitative results, OmniSyn with point cloud rendering achieves worse WS-PSNR results than with mesh rendering. Therefore for outdoor scenes where the depth to the closest regions is small relative to the amount of movement, we suggest using mesh rendering over point cloud rendering to avoid such sparsity issues.

\begin{figure}[ht]
    \centering
    \includegraphics[width=\linewidth]{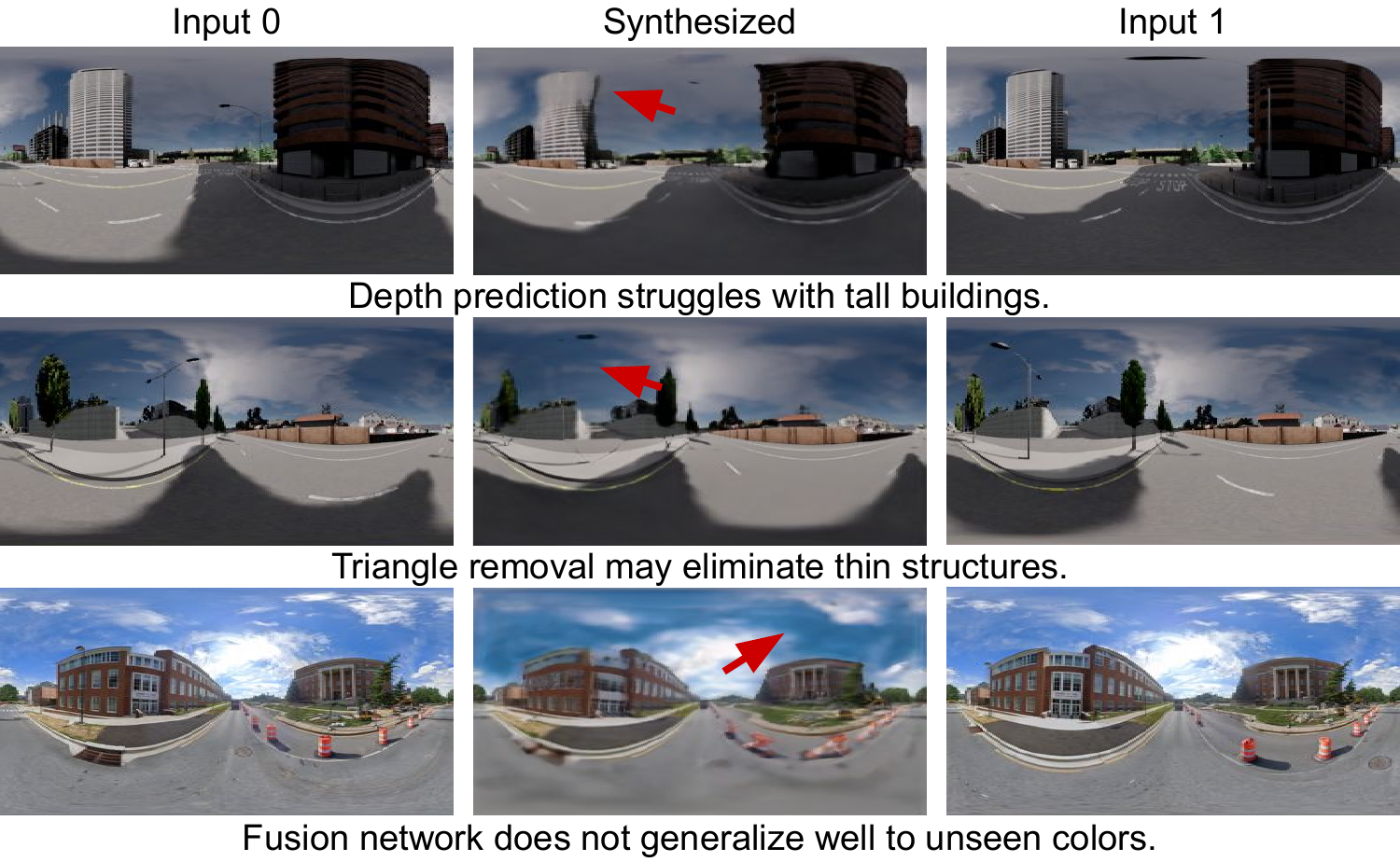}
    \caption{Three types of failure cases which may cause artifacts.}
    \label{fig:failure_cases}
\end{figure}

\subsection{Limitations}

Running our CARLA-trained OmniSyn model on Google Street View images, as shown in \autoref{fig:gsv}, we see that the outputs are reasonable for objects within the CARLA dataset such as buildings and roads. However, thin objects such as wires and tree branches are not well represented in the output. Furthermore, objects such as cars that are not included in the CARLA training procedure are poorly rendered. Similar to the prior art in view synthesis, our pipeline currently focuses on static scenes. While this assumption may hold in street view images of sparsely-populated suburban areas, most real street view images are taken in a dynamic environment filled with moving cars, pedestrians, and objects. 
One way to address this would involve extending object motion detection methods \cite{cao2019motion} to sparse 360\degree street view imagery. High-quality view synthesis for real 360\degree street view scenes continues to remain an open challenge.

We identify three types of failure cases in \autoref{fig:failure_cases}. 
First, spherical sweeping struggles to estimate depth for tall buildings due to distortion in the ERP projection and sweeping levels being more concentrated on closer depths.
Second, triangle culling may lead to thin objects being removed.
Third, the fusion network does not generalize well to unseen objects and colors. One may overcome these limitations by training with a diverse synthetic data which includes moving objects or with large-scale street view datasets. The depth predictor may also be further augmented with rectangular filters~\cite{zioulis2018omnidepth} and a quaternion loss function~\cite{feng2020deep}. 

Our current pipeline and primary results are ran on $256\times256$ resolution images which is unsuitable for high-resolution VR and AR experiences as users only see a limited portion of 360\degree images at any given moment. One of the benefits of using a mesh representation, or similar MPI and LDI representations, is that meshes can be textured with high-resolution images, regardless of the vertex density of the mesh. Therefore, higher resolution results can be achieved by combining a low-resolution depth prediction with a high-resolution mesh rendering as done in previous work \cite{attal2020matryodshka}.
\section{Conclusion}
In this paper, we examine the task of intermediate view synthesis for wide-baseline 360\degree panoramas, typically $\geq5$ meters apart. 
We start by evaluating whether state-of-the-art view synthesis techniques are suitable for creating 360\degree views. 
From our experiments, we observe the following: 
First, current monocular methods hallucinate content that may be inconsistent and unsuitable for wide-baseline $360\degree$ view synthesis. Using stereo images allows the network to more accurately synthesize views and estimate depth. 
Second, for $360\degree$ view synthesis, point cloud renderings incur unnecessary sparsity in nearby objects such as roads. Using mesh rendering better represents the underlying visibility for $360\degree$ images. 
Based on these observations, we develop OmniSyn which leverages 360\degree stereo depth estimation, mesh rendering, and 360\degree fusion to synthesize plausible 360\degree street view panoramas from static scenes.
We envision this line of research may give rise to a wide range of virtual reality applications with depth-based real-time interaction~\cite{Du2020DepthLab}.

\bibliographystyle{abbrv-doi}

\bibliography{template}
\end{document}